\documentclass[11pt]{article}

\pdfoutput=1

\usepackage{emnlp2023}

\usepackage{amsmath}
\usepackage{algorithm}
\usepackage{times}
\usepackage{latexsym}
\usepackage{booktabs}   
\usepackage{graphicx}   

\usepackage{amsmath}
\usepackage{amssymb}
\usepackage{xcolor}
\usepackage{twemojis}
\usepackage[T1]{fontenc}

\usepackage[utf8]{inputenc}

\usepackage{microtype}

\usepackage{inconsolata}
\usepackage{booktabs}
\usepackage{multirow}
\usepackage{comment}

\title{Domain-Agnostic Neural Topic Modeling with Contextual\\ Token-Level Semantic Graph Representation}

\author{
Seung-Won Seo$^{1}$, \quad
Won Ik Cho$^{2}$, \quad
Yongmin Yoo$^{3}$\textsuperscript{$\dagger$} \quad \\
TelePIX$^{1}$, \quad
AI Center, Samsung Electronics$^{2}$ \quad \\
School of Computing \& Frontier AI Research Centre, Macquarie University$^{3}$ \quad \\
\texttt{seungwonseo@telepix.net, \{tsatsuki6,yooyongmin91\}@gmail.com}
}

\begin{document}
\maketitle
{\renewcommand{\thefootnote}{\fnsymbol{footnote}}
\footnotetext[2]{Corresponding author.}} 
\begin{abstract} 

Recent advances in neural topic models with pre-trained language models (PLMs) have achieved strong performance by leveraging general-domain pre-training, yet their topic interpretability often degrades on specialized corpora. This limitation primarily stems from the geometry of the embedding space, where domain-specific terms unseen during pre-training collapse into an indistinguishable region, and neither domain-specific re-training, word-level graph enrichment, nor parameter-efficient fine-tuning can restructure this space without inheriting the capacity ceiling of the underlying encoder. Our key insight is that a learnable graph layer operating on token-level PLM embeddings can acquire corpus-specific semantic structure that the frozen encoder lacks, because token-level graphs preserve document-local context that word-level representations discard and joint optimization with the topic objective reshapes embedding geometry directly from target-domain evidence. We instantiate this insight as \textsc{DARTopic}, a domain-agnostic framework that constructs token-level semantic graphs from frozen PLM embeddings and jointly trains a GNN encoder with topic inference. Across three benchmarks spanning \textit{general}, \textit{biomedical}, and \textit{legal} domains, \textsc{DARTopic} consistently outperforms strong baselines in topic coherence and document clustering without any encoder fine-tuning, while demonstrating robustness to PLM choice and favorable runtime efficiency over fine-tuning-based alternatives.

\end{abstract}
\section{Introduction}

Topic modeling has long been a fundamental technique in natural language processing (NLP), aiming to uncover latent semantic structures from large text collections \citep{wu2024survey}. Topic models have been extensively utilized across various applications, including text generation~\citep{10.1145/3477495.3531990}, sentiment analysis~\citep{9112648} and document summarization~\citep{li-etal-2025-topic}.

Traditional probabilistic approaches such as Latent Dirichlet Allocation (LDA) \citep{blei2003latent}, rely on bag-of-words (BoW) assumptions that disregard word order and fail to capture semantic similarity between lexically distinct but contextually related terms.

To address this, by replacing sparse BoW inputs with dense contextual document embeddings via pre-trained language models (PLMs), recent methods \citep{bianchi-etal-2021-pre, grootendorst2022bertopicneuraltopicmodeling, wu2024fastopic} achieve strong performance, surpassing classical probabilistic approaches~\citep{blei2003latent, hofmann2013probabilistic} and earlier VAE-based neural topic models~\citep{miao2016neural, srivastava2017autoencoding} by a wide margin. However, this success rests on the implicit assumption that the semantic geometry of the PLM embedding space adequately covers the vocabulary and relational structures of the target corpus. When this assumption holds, as in general-domain benchmarks whose distributions closely mirror PLM pre-training data, the resulting topics are coherent and interpretable. 

\begin{table}[t]
    \centering
    \resizebox{0.99\columnwidth}{!}{
    \begin{tabular}{c l}
        \toprule
        \textbf{Model} & \textbf{Top-related words} \\
        \midrule
        \textbf{FASTopic} 
        & develop, immunize, \textcolor{red}{repeat}, elicit, pseudomona, experiment, \\
        & \textcolor{red}{result}, \textcolor{red}{mechanic}, correlate, \textcolor{red}{occur}, \textcolor{red}{feature}, \textcolor{red}{convert} \\
        \midrule
        \textbf{DARTopic} 
        & energy, cerevisiae, oxidative, peroxide, nitrate, hydrogen, \\
        & carbon, kinetics, succinate, respiration, reduction, galactose \\
        \bottomrule
    \end{tabular}}
    \caption{Comparison of topics generated by FASTopic \citep{wu2024fastopic} and \textsc{DARTopic} on BioASQ dataset. We report low-coherence words in \textcolor{red}{RED} font.}
    \label{tab:motivating_examples}
\end{table}

Nevertheless, this assumption may not always hold in specialized domains such as \textit{biomedical} and \textit{legal} domains. Domain-specific terms that appear rarely or never in general pre-training corpora are mapped to a narrow, poorly differentiated region of the embedding space, a pathology known as representation degeneration~\citep{gao2018representation, godey-etal-2024-anisotropy}. For topic modeling, this directly leads the inference network to conflate semantically distinct concepts into incoherent topics when domain-critical terms are represented as geometrically indistinguishable. As illustrated in Table~\ref{tab:motivating_examples}, even FASTopic~\citep{wu2024fastopic}, a recent state-of-the-art model, produces topics on biomedical text that mix domain-relevant terms with generic, low-coherence words such as \textit{repeat}, \textit{mechanic}, and \textit{feature}, confirming that fixed PLM representations fail under domain shift.

Existing remedies each address only a partial aspect of this problem. Domain-specific PLMs such as BioBERT~\citep{lee-etal-2020-biobert} can reshape the embedding geometry but require large-scale pre-training that is unavailable for many specialized fields. Graph-based neural topic models~\citep{xu2023contextguided, adhya-sanyal-2024-ginopic, liu-etal-2025-neural} enrich representations with structural information, yet they operate on word-level graphs with fixed, context-independent node features and therefore cannot capture the contextual variation of the same term across documents. Parameter-efficient fine-tuning via prefix tuning~\citep{li-liang-2021-prefix, akash-chang-2024-enhancing} adapts PLM representations at low cost, but if the frozen encoder cannot geometrically distinguish domain-specific terms in the first place, a lightweight prefix cannot overcome this representational ceiling. None of these approaches resolves the fundamental coupling between topic quality and PLM pre-training coverage.

Our key insight is that this coupling can be broken by interposing a learnable, corpus-specific graph layer between frozen token embeddings and topic inference. Rather than treating PLM outputs as fixed final representations or attempting to adapt the encoder itself, we propose to use them as initial node features for a dynamically constructed token-level semantic graph, which a graph neural network (GNN) then refines jointly with the topic objective. Token-level construction captures contextual semantics that word-level graphs miss, because each token node carries document-specific neighborhood structure. Joint GNN-topic optimization discovers corpus-specific relational patterns that the frozen PLM never observed during pre-training. And because the GNN operates over the full token embedding space, it is not bounded by the narrow capacity of a prefix adapter.

Based on this insight, we propose \textsc{DARTopic} (\textbf{D}omain-\textbf{A}gnostic neu\textbf{R}al \textbf{Topic} modeling), a lightweight framework that constructs token-level semantic graphs from frozen PLM embeddings and jointly trains a GNN encoder with topic inference. As shown in Table~\ref{tab:motivating_examples}, \textsc{DARTopic} generates topics consistently organized around coherent biomedical concepts such as \textit{oxidative stress}, \textit{respiration}, and \textit{carbon metabolism}, achieved without any domain-specific pre-training or fine-tuning.

Our contributions can be summarized as follows:
\begin{itemize}
    \item We show that the performance ceiling of PLM-powered topic models under domain shift arises from fixed embedding geometry, and propose \textsc{DARTopic}, which breaks by leveraging token-level semantic graph representations.

    \item We reveal that token-level semantic graphs capture corpus-specific structure beyond word-level graphs and PLM representations, yielding robust topic quality across various PLMs.

    \item We demonstrate that learning relational structure from target-domain data enables a lightweight graph architecture to bridge domain gaps in frozen PLMs and achieve strong performance across diverse domains.
\end{itemize}
\section{Related Works}

\subsection{PLM-powered Neural Topic Models}
Early neural topic models reformulated topic inference within VAE frameworks~\citep{miao2016neural, srivastava2017autoencoding}, replacing the intractable posteriors of LDA~\citep{blei2003latent} with amortized inference networks, though their reliance on bag-of-words representations limited topic coherence. Subsequent work incorporated static word embeddings such as GloVe~\citep{pennington-etal-2014-glove} to encourage semantically similar words to share topic assignments, partially mitigating this limitation.

The introduction of contextual PLMs marked a more substantial shift: CombinedTM~\citep{bianchi-etal-2021-pre} combines dense Sentence-BERT~\citep{reimers-gurevych-2019-sentence} embeddings with BoW features, while BERTopic~\citep{grootendorst2022bertopicneuraltopicmodeling} adopts a clustering-based approach leveraging contextual embeddings to group semantically coherent documents. More recent models such as FASTopic~\citep{wu2024fastopic} and NeuroMax~\citep{pham-etal-2024-neuromax} further advance this paradigm through optimal transport-based alignment and mutual information maximization, respectively, achieving strong performance on general-domain benchmarks.

Despite their effectiveness, PLM-powered models are fundamentally constrained by their pre-training domain coverage. When applied to specialized domains such as biomedical or legal text, the divergence between pre-training and target distributions leads to degraded topic coherence. Domain-specific PLMs~\citep{lee-etal-2020-biobert} can partially address this gap but require costly large-scale pre-training that is not always feasible. PVTM~\citep{akash-chang-2024-enhancing} mitigates domain shift via prefix tuning~\citep{li-liang-2021-prefix} of the PLM encoder, yet its quality remains tightly coupled to the underlying PLM capacity. In particular, lightweight models with limited prior knowledge struggle to produce coherent topics under significant domain shift, regardless of the tuning strategy applied.

\subsection{Graph-based Neural Topic Models}

Graph neural networks (GNNs) have demonstrated strong representational capacity across NLP tasks~\citep{wu2023graph}, and several lines of work have explored their application to topic modeling. Early graph-based models focused on alleviating data sparsity in short-text settings, where limited token co-occurrence information impedes reliable topic inference~\citep{6778764, zhu-etal-2018-graphbtm, wang-etal-2021-extracting}, enriching structural signals via corpus-level word co-occurrence graphs. Zhou et al.~\citeyearpar{zhou-etal-2020-neural} extended this by explicitly modeling document-level relational structures at the corpus level.

More recent work integrates GNNs with pre-trained embeddings: GINopic~\citep{adhya-sanyal-2024-ginopic} employs graph isomorphism network (GIN) over word graphs to capture higher-order relational structures, CGTM~\citep{liu-etal-2025-neural} fuses PLM-based contextual document embeddings with graph-structured word representations, and Meta-CETM~\citep{xu2023contextguided} uses a variational graph auto-encoder over dependency-parsed word graphs for low-resource settings. However, all of these approaches rely on fixed pre-trained embeddings, propagating domain-mismatched semantics through the graph regardless of architecture.

Taken together, PLM and graph-based topic models share a fundamental vulnerability: their representations are anchored to fixed pre-training knowledge that may not reflect the semantic structures of the target domain. \textsc{DARTopic} addresses this gap by constructing token-level semantic graphs whose node representations are initialized from a frozen PLM and jointly refined with the topic inference objective, learning corpus-adaptive representations directly from target data without reliance on domain-specific pre-training.
\section{Proposed Methodology}

\begin{figure*}[ht]
    \centering
    \includegraphics[width=0.99\textwidth]{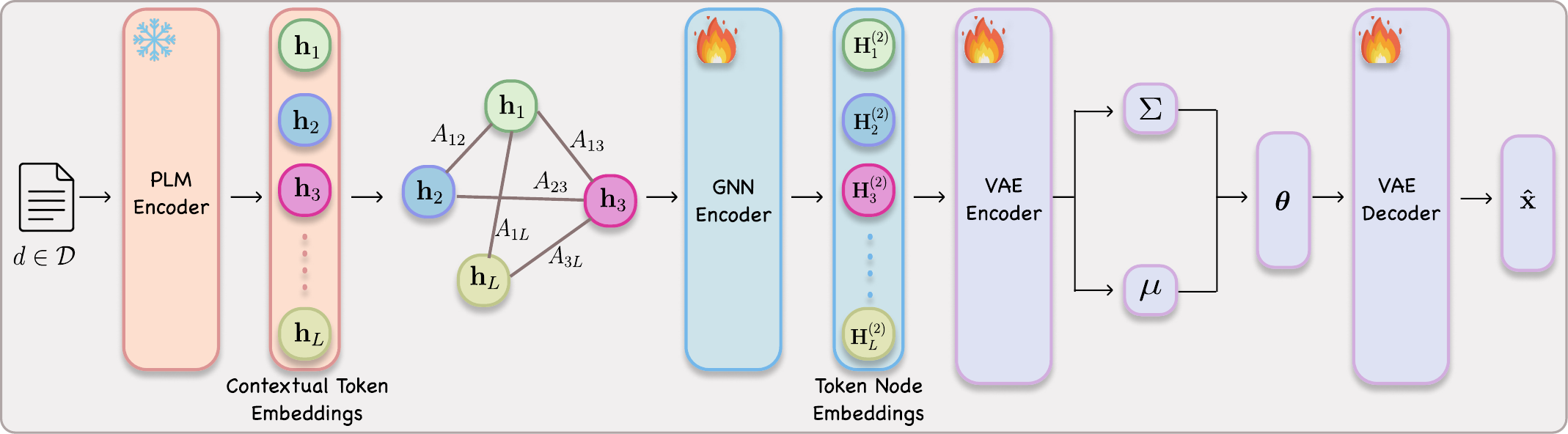}
    \caption{The overall architecture of our proposed \textsc{DARTopic} framework. \twemoji{snowflake} denotes frozen parameters, while \twemoji{fire} denotes trainable parameters.}
    \label{fig:DARTopic_framework}
\end{figure*}

\subsection{Problem Statement}

\textbf{Notations.} Let $\mathcal{D} = \{d_1, d_2, ..., d_N\}$ denote a corpus consisting of $N$ documents. Each document $d_i$ is represented as a sequence of tokens $d_i = (t_{i1}, t_{i2}, ..., t_{iL_i})$, where $L_i$ denotes the number of tokens in the document and each token $t_{ij}$ corresponds to a word from a vocabulary set $\mathcal{V}$ with size $V = |\mathcal{V}|$. In topic modeling, documents are commonly represented using a bag-of-words (BoW) representation. 
Let $\mathbf{x}_i \in \mathbb{R}^{V}$ denote the BoW vector for document $d_i$, where $\mathbf{x}_{iv}$ indicates the frequency of word $v$ in the document. A topic model assumes the existence of $K$ latent topics. Each document is associated with a document-topic distribution $\boldsymbol{\theta}_i \in \Delta^{K}$, where $\Delta^{K}$ denotes a $K$-dimensional probability simplex. Each topic $k$ is characterized by a topic-word distribution $\boldsymbol{\beta}_k \in \Delta^{V}$, and the topic-word matrix is denoted as $\boldsymbol{\beta} \in \mathbb{R}^{K \times V}$ where $\beta_{kv}$ represents the probability of word $v$ under topic $k$. The log-likelihood of document $d_i$ under the topic model is given by \citep{blei2003latent}:
\begin{equation}
\log p(d_i \mid \theta_i, \beta)
=
\sum_{v=1}^{V}
x_{iv}
\log
\left(
\sum_{k=1}^{K} \theta_{ik} \beta_{kv}
\right)
\end{equation}

\textbf{Topic Modeling with Document Representations.} Recent neural topic models often incorporate pre-trained language encoders to obtain semantic document representations \citep{bianchi-etal-2021-cross, bianchi-etal-2021-pre, pham-etal-2024-neuromax, wu2024fastopic, akash-chang-2024-enhancing, liu-etal-2025-neural}. 
Let $f(\cdot)$ denote a pre-trained encoder that maps a document $d_i$ into a dense vector representation $\mathbf{z}_i \in \mathbb{R}^{H}$, where $H$ denotes the embedding dimension, i.e., $\mathbf{z}_i = f(d_i)$. Such encoders are typically trained on large-scale corpora and provide contextualized semantic representations of documents. 
In neural topic modeling frameworks, the document representation $\mathbf{z}_i$ is used as input to an inference network that estimates the document-topic distribution. 
Formally, the document-topic distribution is obtained as $\boldsymbol{\theta}_i = g(\mathbf{z}_i)$, where $g(\cdot)$ denotes a inference network. The topic model then learns topic-word distributions $\boldsymbol{\beta}$ by maximizing the log-likelihood of documents given the inferred topic mixtures:

\begin{equation}
\max_{\beta, g}
\sum_{i=1}^{N}
\log p(d_i \mid \theta_i, \beta),
\quad
\text{where } \theta_i = g(f(d_i)).
\end{equation}

Since contextualized document representations obtained from a pre-trained encoder often capture rich semantic information, they can significantly contribute to the discovery of coherent topics. Such contextual information can help the topic model better identify semantically related words and improve the overall quality of the learned topic-word distributions. Therefore, our main goal is to effectively leverage contextual information derived from the pre-trained encoder while adapting it to heterogeneous domain corpora.


\subsection{Token-Level Semantic Graph Construction}

To effectively represent documents from heterogeneous domains, we construct a token-level graph representation by leveraging token embeddings obtained from a pre-trained language model (PLM). Inspired by \citep{angelov-inkpen-2024-topic, donabauer2025token}, we represent each document as a graph where nodes correspond to tokens and edges capture relationships between tokens within the same document.

The motivation for adopting a token-level graph representation is closely related to the domain-agnostic setting considered in this work. When documents originate from diverse domains, contextual information available in the training corpus may be insufficient to learn high-quality word representations without extensive fine-tuning. In such scenarios, relying solely on corpus-level statistics can lead to suboptimal semantic representations.

To alleviate this issue, we explicitly leverage the knowledge embedded in a PLM by constructing graphs based on token embeddings extracted from the PLM. This design allows the model to incorporate rich semantic knowledge acquired during large-scale pretraining while maintaining robustness under low-resource or cross-domain conditions. Unlike the token-level graph construction strategy \citep{donabauer2025token}, which connects tokens based on local co-occurrence windows (i.e., an $n$-hop or sliding-window graph), we construct a semantic graph where edges are determined by the similarity between token embeddings. The difference in graph construction reflects the different inductive biases required by the target task. Local window-based graphs emphasize positional proximity and syntactic relations, which are beneficial for text classification tasks \citep{wang-etal-2021-cross-lingual, wang2024graph}. In contrast, topic modeling aims to uncover latent semantic structures among words across a document. Therefore, a graph that directly captures semantic similarity provides a more appropriate inductive bias for topic modeling task. A detailed empirical analysis comparing different graph construction strategies is presented in Appendix \ref{sec:graph_contruction}. 

To obtain token embeddings, we employ the lightweight small-scale pre-trained model\footnote{\url{https://huggingface.co/sentence-transformers/all-MiniLM-L6-v2}}, which offers a good trade-off between computational efficiency and representation quality, making it suitable for low-resource settings. The tokenizer used in our framework is defined as \texttt{WordPiece} tokenizer \citep{song-etal-2021-fast}. Given a document containing $L$ tokens, we denote the token embedding matrix as $\mathbf{H} = [\mathbf{h}_1, \mathbf{h}_2, \dots, \mathbf{h}_L] \in \mathbb{R}^{L \times d}$, where $\mathbf{h}_i \in \mathbb{R}^d$ represents the contextual embedding of the $i$-th token obtained from the PLM. We construct an undirected and weighted token-level semantic graph $G = (V, E)$, where each node $v_i \in V$ corresponds to a token embedding $\mathbf{h}_i$. The edges are defined based on cosine similarity between token embeddings. Specifically, the adjacency matrix $\mathbf{A} \in \mathbb{R}^{L \times L}$ is defined as follows:

\begin{equation}
A_{ij} =
\begin{cases}
\cos(\mathbf{h}_i, \mathbf{h}_j), & \text{if } \cos(\mathbf{h}_i, \mathbf{h}_j) \ge \tau \\
0, & \text{otherwise}
\end{cases}
\end{equation}

where $\cos(\mathbf{h}_i, \mathbf{h}_j)$ denotes the cosine similarity between token embeddings and $\tau$ is a predefined similarity threshold controlling graph sparsity. The constructed token-level semantic graph is subsequently processed by a graph neural network to learn contextualized node representations, which are then used for topic inference.

\subsection{Graph Representation Learning}
We employ a two-layer graph convolutional network (GCN)~\citep{kipf2017semisupervised} to propagate semantic information over the token graph and obtain node embeddings that encode both token semantics and document-specific structural relations.

Let a document graph be denoted as $G = (V, E)$ with adjacency matrix $\mathbf{A} \in \mathbb{R}^{L \times L}$ and node feature matrix $\mathbf{X} \in \mathbb{R}^{L \times d}$, where $L$ is the number of tokens in the document and $d$ is the dimensionality of the PLM-based token embeddings. Following the standard GCN formulation~\citep{kipf2017semisupervised}, we apply symmetric normalization with self-loops: $\hat{\mathbf{A}} = \tilde{\mathbf{D}}^{-\frac{1}{2}} (\mathbf{A} + \mathbf{I}) \tilde{\mathbf{D}}^{-\frac{1}{2}}$. The first layer computes hidden representations as $\mathbf{H}^{(1)} = \mathrm{ReLU}(\hat{\mathbf{A}}\mathbf{X}\mathbf{W}^{(1)} + \mathbf{b}^{(1)})$, and the second layer produces the final node representations as

\begin{equation}
\mathbf{H}^{(2)} = \mathrm{ReLU}(\hat{\mathbf{A}}\mathbf{H}^{(1)}\mathbf{W}^{(2)} + \mathbf{b}^{(2)}),
\end{equation}

where $\mathbf{W}^{(\cdot)}$ and $\mathbf{b}^{(\cdot)}$ are learnable parameters of each layer. The resulting matrix $\mathbf{H}^{(2)} \in \mathbb{R}^{L \times d'}$ contains the contextualized embeddings of all token nodes. Graph convolution enables each token representation to be enriched by semantically related neighbors in the document graph, which is particularly desirable for topic modeling where semantically coherent groups of tokens should be aggregated into document-level latent representations. To derive a fixed-length document representation, we apply mean pooling over all token nodes:

\begin{equation}
\label{eq:6}
\mathbf{g} = \frac{1}{L} \sum_{i=1}^{L} \mathbf{H}^{(2)}_{i},
\end{equation}

where $\mathbf{H}^{(2)}_{i}$ denotes the contextualized embedding of the $i$-th token node.


\subsection{\textsc{DARTopic} Framework}

In this subsection, we present the overall \textsc{DARTopic} framework. The overall architecture is illustrated in Figure~\ref{fig:DARTopic_framework}, our framework is designed to remain lightweight while effectively leveraging PLM-derived token semantics for neural topic modeling on various domains. To this end, \textsc{DARTopic} consists of a graph encoder for token-level document representation learning and a compact VAE-based topic inference module.

\textbf{Lightweight Architecture.} 
A key design principle of \textsc{DARTopic} is to keep the overall architecture lightweight. Unlike prior contextualized neural topic models that use dense contextual representations together with sparse lexical features such as TF-IDF or bag-of-words (BoW) vectors \citep{bianchi-etal-2021-cross, adhya-sanyal-2024-ginopic}, we do not employ any joint input structure with sparse document embeddings. Such hybrid architectures often require additional projection layers or high-dimensional transformations, which increase the number of learnable parameters. Instead, we use only the graph-level document representation $\mathbf{g} \in \mathbb{R}^{d'}$ obtained by mean pooling over all token nodes, as defined in Eq.\ref{eq:6}. By directly using $\mathbf{g}$ as the input of the topic model, \textsc{DARTopic} avoids the need to concatenate additional sparse document features, reducing computational cost and minimizing the number of trainable parameters.

\textbf{Topic Inference.} 
Following~\citep{bianchi-etal-2021-cross}, we employ a variational autoencoder (VAE) to infer the latent topic structure of the documents. The encoder maps $\mathbf{g}$ to the parameters of a Gaussian posterior distribution over the latent variable $\mathbf{z} \in \mathbb{R}^{K}$, where $K$ denotes the number of topics. Concretely, the encoder produces the mean vector $\boldsymbol{\mu}$ and log-variance vector $\log \boldsymbol{\sigma}^2$ as $\boldsymbol{\mu} = f_{\mu}(\mathbf{g})$ and $\log \boldsymbol{\sigma}^2 = f_{\sigma}(\mathbf{g})$, where $f_{\mu}(\cdot)$ and $f_{\sigma}(\cdot)$ denote neural transformations of VAE. To enable gradient-based optimization, we adopt the reparameterization trick \citep{kingma2013auto} and sample the latent variable as $\mathbf{z} = \boldsymbol{\mu} + \boldsymbol{\sigma} \odot \boldsymbol{\epsilon}$, where $\boldsymbol{\epsilon} \sim \mathcal{N}(\mathbf{0}, \mathbf{I})$ and $\odot$ denotes element-wise multiplication. The sampled latent variable is then transformed into the document-topic distribution $\boldsymbol{\theta} \in \mathbb{R}^{K}$ through a softmax function, i.e., $\boldsymbol{\theta} = \mathrm{softmax}(\mathbf{z})$.

\textbf{Overall Training Objective.} 
The GNN encoder and the VAE-based neural topic model are trained jointly in an end-to-end unsupervised manner. Let $\mathbf{x}_{\mathrm{bow}} \in \mathbb{R}^{V}$ denote the target document word-count vector over a vocabulary set of size $V$, and let $\hat{\mathbf{x}} \in \mathbb{R}^{V}$ denote the reconstructed word distribution generated from the inferred topic mixture $\boldsymbol{\theta}$ through the decoder. The reconstruction loss is defined as the negative log-likelihood of the observed document under the reconstructed word distribution:

\begin{equation}
\label{eq:8}
\begin{aligned}
\mathcal{L}_{\mathrm{rec}} &= \frac{1}{N} \sum_{i=1}^{N} 
\left[ - (\mathbf{x}_{i})^\top \log \left( \text{softmax} (\boldsymbol{\beta \theta_i)} \right) \right]. \\
\end{aligned}
\end{equation}

To regularize the latent topic space, we impose a standard Gaussian prior on $\mathbf{z}$, i.e., $p(\mathbf{z}) = \mathcal{N}(\mathbf{0}, \mathbf{I})$. The Kullback--Leibler (KL) divergence between the approximate posterior $q(\mathbf{z} \mid \mathbf{g})$ and the prior is defined as follows:
\begin{equation}
\mathcal{L}_{\mathrm{KL}}
=
D_{\mathrm{KL}}\bigl(q(\mathbf{z}\mid\mathbf{g}) \,\|\, p(\mathbf{z})\bigr).
\end{equation}

The overall training objective of \textsc{DARTopic} is given by $\mathcal{L} = \mathcal{L}_{\mathrm{rec}} + \mathcal{L}_{\mathrm{KL}}$. By optimizing this objective, the GCN-based graph encoder and the VAE-based topic model are trained simultaneously. Importantly, the token-level graph representation learning module is not optimized with a separate graph-specific supervision signal; instead, it is directly guided by the neural topic modeling objective. This allows the learned token and document representations to be specialized for latent topic inference while preserving the lightweight nature of the overall framework.

\begin{table*}[t]
    \centering
    \resizebox{\textwidth}{!}{
    \begin{tabular}{lllcccccccccc}
        \toprule
        Dataset & Domain & Metric
        & ProdLDA & FASTopic & BERTopic & ZeroShotTM & CombinedTM
        & NeuroMax & PVTM & GINopic & CGTM & \textsc{DARTopic} \textit{(ours)} \\
        \midrule

        \multirow{3}{*}{20NG}
        & \multirow{3}{*}{\textit{General}}
        & NPMI & 0.1581 & 0.2624 & 0.2300 & 0.2604 & 0.1670 & 0.1833 & \underline{0.2707} & 0.2172 & 0.2520 & \textbf{0.2736} \\
        & & TU          & 0.7613 & 0.8813  & 0.4547  & 0.9227  & 0.7853 & \textbf{1.0000} & 0.9133 & 0.4747 & 0.5227 & \underline{0.9360} \\
        & & TQ          & 0.1203 & 0.2313 & 0.1046 & 0.2403 & 0.1311 & 0.1833 & \underline{0.2473} & 0.1031 & 0.1317 & \textbf{0.2561} \\
        \midrule

        \multirow{3}{*}{BioASQ}
        & \multirow{3}{*}{\textit{Biomedical}}
        & NPMI   & 0.0142 & 0.0789 & 0.1653 & 0.1448 & 0.0120 & 0.0094 & \underline{0.1596} & 0.0213 & 0.1377 & \textbf{0.1641} \\
        & & TU          & 0.8213 & 0.4093 & 0.7320 & 0.8640 & 0.8080 & \textbf{1.0000} & 0.9730 & 0.8013 & 0.6947 & \underline{0.9760} \\
        & & TQ          & 0.0117 & 0.0323 & 0.1210 & 0.1251 & 0.0097 & 0.0094 & \underline{0.1553} & 0.0171 & 0.0957 & \textbf{0.1602} \\
        \midrule

        \multirow{3}{*}{Bills}
        & \multirow{3}{*}{\textit{Legal}}
        & NPMI   & 0.0387 & 0.2291 & 0.1795 & 0.2525 & 0.0425 & 0.0506 & \underline{0.2609} & 0.0371 & 0.2318 & \textbf{0.2685} \\
        & & TU          & 0.7853 & 0.9720 & 0.4787 & 0.9000 & 0.7813 & \textbf{1.0000} & 0.9133 & 0.8080 & 0.6013 & \underline{0.9240} \\
        & & TQ          & 0.0304 & 0.2226 & 0.0859 & 0.2272 & 0.0332 & 0.0506 & \underline{0.2383} & 0.0300 & 0.1394 & \textbf{0.2481} \\
        \bottomrule
    \end{tabular}}
    \caption{Main results on topic-word distribution quality across three domains: \textit{general} (20NG), \textit{biomedical} (BioASQ) and \textit{legal} (Bills). The best performance in each row is shown in \textbf{bold} and the second-best is \underline{underlined}.}
    \label{tab:npmi-4datasets-acl-domain}
\end{table*} 

\section{Experiments}
\subsection{Experimental Setup}
\label{sec:experimental setup}

\textbf{Datasets.} We conduct experiments on three benchmarks with heterogeneous domain texts: 20NewsGroup (20NG) (\textit{General}), BioASQ (\textit{Biomedical}) and BillSum (Bills) \citep{kornilova2019billsum} (\textit{Legal}). The statistics of the processed datasets are shown in Table~\ref{tab:dataset_statistics} in Appendix~\ref{sec:appendix dataset}.
\smallskip\\
\textbf{Baselines.} We compare our framework with the following neural topic models: (1) \textbf{FASTopic} \citep{wu2024fastopic}, the recent state-of-the-art neural topic model using embedding regularization with optimal transport, (2) \textbf{BERTopic} \citep{grootendorst2022bertopicneuraltopicmodeling}, a standard clustering-based topic model, 
(3) \textbf{ProdLDA} \citep{srivastava2017autoencoding}, a first VAE-based standard neural topic model, (4) \textbf{ZeroShotTM} \citep{bianchi-etal-2021-cross} which replaces input data with PLM-based document representation, (5) \textbf{CombinedTM}~\citep{bianchi-etal-2021-pre} which integrates BoW embeddings with PLM representations, (6) \textbf{PVTM} \citep{akash-chang-2024-enhancing}, a VAE-based neural topic model with parameter-efficient fine-tuning method, (7) \textbf{GINopic} \citep{adhya-sanyal-2024-ginopic}, a graph-based neural topic model with graph isomorphism network, (8) \textbf{CGTM} \citep{liu-etal-2025-neural}, a neural topic model utilizing both Graph and PLMs, (9) \textbf{NeuroMax} \citep{pham-etal-2024-neuromax}, a lightweight neural topic model using PLM only in training time. The implementation details of baseline models are shown in Appendix \ref{sec:implementation}.
\smallskip\\
\textbf{Evaluation.} 
We use automatic evaluation metrics such as NPMI~\citep{lau-etal-2014-machine} (topic coherence) and TU~\citep{nan-etal-2019-topic} (topic diversity). For topic coherence evaluation on domain-specific texts, we follow~\citep{han-etal-2023-unified} and use the NPMI-In score. To evaluate coherence and diversity as a whole, we also use Topic Quality (TQ)~\citep{dieng-etal-2020-topic}. 
Furthermore, to evaluate the quality of document-topic distributions, we conduct document clustering task and report Purity and normalized mutual information (NMI) \citep{Manning_IR_2008}. NMI measures the agreement between predicted topic assignments and ground-truth labels, while Purity assesses how consistently each cluster contains documents from a single class. Please refer to Appendix~\ref{sec:evaluation_metrics} for the detailed.

\subsection{Main Results}
\paragraph{Topic-Word Distribution Quality.}

We report the mean value over 5 random runs with 50 topics. Table~\ref{tab:npmi-4datasets-acl-domain} reports the topic-word distribution quality results on three datasets with different domains. Overall, \textsc{DARTopic} achieves consistently strong performance across all datasets, demonstrating its robustness to both domain variation and document length. In particular, the average document lengths of the three corpora are substantially different, i.e., 48.02 for 20NG, 7.44 for BioASQ, and 76.28 for Bills, yet \textsc{DARTopic} maintains superior or highly competitive results in all cases. This indicates that the proposed framework can effectively capture topic structure regardless of whether the target corpus consists of short or long documents. PVTM shows strong overall performance by leveraging parameter-efficient fine-tuning (PEFT) such as prefix tuning \citep{li-liang-2021-prefix} to adapt pretrained language models to domain-specific corpora. NeuroMax, in contrast, achieves perfect or near-perfect topic diversity through its embedding clustering regularization (ECR) \citep{wu2023effective}, but this comes at the cost of substantially weaker topic coherence. Compared with these methods, \textsc{DARTopic} provides a more balanced and reliable trade-off between coherence and diversity. Specifically, \textsc{DARTopic} achieves the best NPMI and TQ scores on all three datasets, while also maintaining highly competitive TU scores, demonstrating that it can generate topics that are not only diverse but also semantically coherent. These results suggest that \textsc{DARTopic} is not merely specialized for a particular domain or corpus condition, but instead offers a generally effective topic modeling framework across heterogeneous datasets. Its consistent gains over strong baselines further highlight the advantage of graph-based document representation learning for improving topic quality in neural topic modeling.

\textbf{Doc-Topic Distribution Quality.}
We evaluate document-topic distributions via a downstream clustering task. The Bills dataset lacks document labels and is also omitted. Following~\citet{adhya-sanyal-2024-ginopic}, each document is assigned to its highest-probability topic. Table~\ref{tab:doc-topic distribution of topic_models} reports the results. \textsc{DARTopic} shows strong performance on both 20NG and BioASQ, with a clearer advantage on BioASQ, where domain-specific terms are poorly separated in the frozen PLM space and the graph layer's corpus-learned structure provides greater discriminative benefit.

\begin{table}[ht] 
    \centering
    \resizebox{0.99\linewidth}{!}{%
    \begin{tabular}{lccccc}
        \toprule
        & NPMI & TU & TQ & Purity & NMI \\
        \midrule
        \texttt{MiniLM} & 0.1641 & 0.9760 & 0.1602 & 0.5235 & 0.3833 \\
        \texttt{Roberta-large} & 0.1660 & 0.9680 & 0.1606 & 0.4738 & 0.3492 \\
        \texttt{Qwen-0.6B} & 0.1644 & 0.9653 & 0.1587 & 0.4829 & 0.3555 \\ 
        \texttt{BioBERT} & 0.1690 & 0.9693 & 0.1638 & 0.5061 & 0.3696 \\
        \bottomrule
    \end{tabular}%
    }
    \caption{Comparison of \textsc{DARTopic} performance on BioASQ dataset with different pre-trained models.}
    \label{tab:robustness_analysis_DARTopic}
\end{table}

\begin{table}[h]
    \centering
    \resizebox{0.95\columnwidth}{!}{ 
    \begin{tabular}{lcc|cc}
        \toprule
        & \multicolumn{2}{c}{20NG} & \multicolumn{2}{c}{BioASQ} \\
        \cmidrule(lr){2-3} \cmidrule(lr){4-5}
        & Purity & NMI & Purity & NMI \\
        \midrule
        ProdLDA & 0.1186 & 0.1652 & 0.2032 & 0.1030 \\
        BERTopic & 0.4347 & 0.3369 & 0.4666 & 0.3205 \\
        FASTopic & 0.4743 & \textbf{0.4089} & 0.4339 & 0.3650 \\
        ZeroshotTM & 0.4555 & 0.3657 & 0.3217 & 0.3312 \\
        CombinedTM & 0.1963 & 0.2262 & 0.2466 & 0.1566 \\
        NeuroMax & 0.1889 & 0.2530 & 0.2561 & 0.2802 \\
        PVTM & \textbf{0.4927} & 0.3917 & \underline{0.5174} & \underline{0.3824} \\
        GINopic & 0.2122 & 0.2523 & 0.2440 & 0.1619 \\
        CGTM & 0.4562 & 0.3900 & 0.4809 & 0.3626 \\
        \midrule
        \textsc{DARTopic} \textit{(ours)} & \underline{0.4905} & \underline{0.3968} & \textbf{0.5235} & \textbf{0.3833} \\
        \bottomrule
    \end{tabular}}
    \caption{Performance comparison on document clustering task using document-topic distribution vector. The best-performing method is highlighted in \textbf{bold} and second-best performance is \underline{underlined}.}
    \label{tab:doc-topic distribution of topic_models} 
\end{table}

\section{Analysis}

\begin{table}[t] 
    \centering
    \resizebox{\columnwidth}{!}{%
    \begin{tabular}{lcccccc}
        \toprule
        Methods & PLMs & NPMI & TU & TQ & Purity & NMI  \\
        \midrule
        FASTopic & \texttt{MiniLM} & 0.0789 & 0.4093 & 0.0323 & 0.4339 & \textbf{0.3650} \\
        FASTopic & \texttt{BioBERT} & \textbf{0.1146} & \textbf{0.9920} & \textbf{0.1137} & \textbf{0.4439} & 0.3375 \\
        \midrule
        ZeroshotTM & \texttt{MiniLM} & 0.1448 & 0.8640 & 0.1251 & 0.3217 & 0.3312 \\
        ZeroshotTM & \texttt{BioBERT} & \textbf{0.1628} & \textbf{0.9680} & \textbf{0.1576} & \textbf{0.4967} & \textbf{0.3652} \\
        \midrule
        CombinedTM & \texttt{MiniLM} & 0.0120 & \textbf{0.8080} & 0.0097 & \textbf{0.2466} & \textbf{0.1566} \\
        CombinedTM & \texttt{BioBERT} & \textbf{0.0130} & 0.7813 & \textbf{0.0102} & 0.1108 & 0.1157 \\
        \midrule
        BERTopic & \texttt{MiniLM} & \textbf{0.1653} & \textbf{0.7320} & \textbf{0.1210} & \textbf{0.4666} & \textbf{0.3205} \\
        BERTopic & \texttt{BioBERT} & 0.1601 & 0.7013 & 0.1123 & 0.4247 & 0.3031 \\
        \midrule
        NeuroMax & \texttt{MiniLM} & \textbf{0.0094} & \textbf{1.0000} & \textbf{0.0094} & \textbf{0.2561} & \textbf{0.2802} \\
        NeuroMax & \texttt{BioBERT} & 0.0073 & \textbf{1.0000} & 0.0073 & 0.2490 & 0.1994 \\
        \midrule
        PVTM & \texttt{MiniLM} & 0.1596 & \textbf{0.9730} & 0.1553 & \textbf{0.5174} & \textbf{0.3824} \\
        PVTM & \texttt{BioBERT} & \textbf{0.1716} & 0.9707 & \textbf{0.1666} & 0.4908 & 0.3651 \\
        \midrule
        CGTM & \texttt{MiniLM} & \textbf{0.1377} & \textbf{0.6947} & \textbf{0.0957} & \textbf{0.4809} & \textbf{0.3626} \\
        CGTM & \texttt{BioBERT} & 0.1366 & 0.6040 & 0.0825 & 0.4632 & 0.3437 \\
        \bottomrule
    \end{tabular}
    }
    \caption{PLM dependency of PLM-powered neural topic models performance on BioASQ dataset. The better result is highlighted in \textbf{bold} for each metric.}
    \label{tab:robustness_analysis_baselines}
\end{table}

\subsection{Robustness Analysis}

We analyze the robustness of neural topic models under different PLM choices, including general-purpose PLMs and a domain-specific PLM, \texttt{BioBERT}. 
Tables~\ref{tab:robustness_analysis_DARTopic} and~\ref{tab:robustness_analysis_baselines} report the results on the BioASQ dataset.

Table~\ref{tab:robustness_analysis_DARTopic} shows the performance of \textsc{DARTopic} with various PLMs, ranging from lightweight models such as \texttt{MiniLM} to larger or domain-specific encoders such as \texttt{RoBERTa-large}, \texttt{Qwen-0.6B}, and \texttt{BioBERT}. 
The results demonstrate that \textsc{DARTopic} achieves consistently strong performance across different PLM choices, with only limited variation. 
This indicates that \textsc{DARTopic} is robust to both the scale and type of the underlying encoder, making it practical even when smaller and more efficient PLMs are preferred.

Table~\ref{tab:robustness_analysis_baselines} further compares recent PLM-based neural topic models using \texttt{MiniLM} and \texttt{BioBERT}. 
The results show that existing methods exhibit different levels of sensitivity to PLM choice. 
For models that combine PLM-based document embeddings with additional representations such as BoW or word embeddings, including CombinedTM, CGTM, and NeuroMax, \texttt{MiniLM} often outperforms \texttt{BioBERT}. 
In contrast, PLM-driven models such as FASTopic and ZeroShotTM benefit substantially from replacing \texttt{MiniLM} with \texttt{BioBERT}. 

Overall, these results suggest that many baseline methods are highly dependent on the selected PLM, whereas \textsc{DARTopic} maintains stable and competitive performance regardless of the encoder. 
This demonstrates that the proposed graph-based framework can preserve topic quality even with relatively small or generic PLMs, offering a more robust solution for practical neural topic modeling.

\begin{table}[h]
    \centering
    \resizebox{0.99\columnwidth}{!}{
    \begin{tabular}{lcc|cc|cc}
        \toprule
        & \multicolumn{2}{c}{20NG} & \multicolumn{2}{c}{BioASQ} & \multicolumn{2}{c}{Bills} \\
        \cmidrule(lr){2-3} \cmidrule(lr){4-5} \cmidrule(lr){6-7}
        & Training & Inference & Training & Inference & Training & Inference \\
        \midrule
        PVTM & 13.55s & 12.79s & 2.55s & 2.22s & 9.82s & 9.56s \\
        DARTopic & 7.67s & 7.65s & 2.59s & 2.44s & 8.67s & 8.52s \\
        \bottomrule
    \end{tabular}}
    \caption{Runtime performance comparison with PVTM.}
    \label{tab:runtime}
\end{table}

\subsection{Runtime Performance}
We compare PVTM and \textsc{DARTopic} in terms of per-epoch training time and full-corpus inference time, where all experiments were conducted on a single NVIDIA H100 GPU. In Table~\ref{tab:runtime}, the results show that \textsc{DARTopic} is consistently faster than PVTM on 20NG and Bills, while maintaining comparable efficiency on BioASQ. Notably, the advantage of \textsc{DARTopic} becomes more pronounced on datasets with longer documents, such as 20NG and Bills. This trend suggests that \textsc{DARTopic} is particularly effective for long-document processing, where computational bottlenecks tend to increase with document length. While PVTM relies on Prefix Tuning to mitigate domain mismatch, such adaptation introduces additional trainable parameters and computational overhead. In contrast, \textsc{DARTopic} uses a frozen pretrained encoder, enabling a simpler and more efficient architecture. As a result, \textsc{DARTopic} reduces the runtime burden associated with increasing document length, demonstrating that efficient domain-agnostic topic modeling can be achieved without fine-tuning. Notably, as shown in Table~\ref{tab:npmi-4datasets-acl-domain}, \textsc{DARTopic} achieves this efficiency while also outperforming PVTM in topic quality across all three datasets.

\subsection{Summary of Key Findings}

Based on our experiments and analysis, we highlight three key findings:
\medskip\\
\textbf{Decoupling Topic Quality from PLM Capacity} \textsc{DARTopic} achieves the best NPMI and TQ across all three domains using only a frozen PLM, demonstrating that corpus-learned token-level graph representations can replace domain-specific pre-training.\smallskip\\
\textbf{Robustness to PLM Choice.} Unlike existing models that are sensitive to encoder selection, \textsc{DARTopic} maintains stable performance across PLMs of varying scale and domain specificity.
\smallskip\\
\textbf{Efficiency without Fine-Tuning.} Compared to PVTM which relies on prefix tuning, \textsc{DARTopic} achieves superior topic quality while being faster, confirming that graph representation learning is a more effective alternative to fine-tuning for cross-domain topic modeling.
\section{Conclusion}

We introduced \textsc{DARTopic}, which decouples topic quality from PLM pre-training coverage by interposing a corpus-learned token-level semantic graph between frozen embeddings and variational topic inference. Experiments across general, biomedical, and legal domains confirm consistent improvements in topic coherence and document clustering, with robustness to PLM choice and lower runtime than fine-tuning alternatives. Our results suggest a broader principle: when a frozen encoder lacks domain-specific geometric structure, a jointly optimized graph layer can reconstruct it from target-corpus evidence alone.
\section*{Limitations}

Our proposed \textsc{DARTopic} framework shows strong effectiveness across heterogeneous domains, consistently improving topic quality and document representations with an efficient small-scale PLM-powered design. This work has several limitations that should be addressed in future research. Although \textsc{DARTopic} shows consistent gains across \textit{general}, \textit{biomedical}, and \textit{legal} benchmarks, our experiments are limited to English corpora. Because the framework relies on token-level semantic graph construction rather than domain-specific resources, it is potentially applicable to broader settings, but multilingual validation remains for future work. In addition, our evaluation mainly uses standard automatic topic coherence metrics. While these protocols are widely adopted in neural topic modeling, human evaluation could provide complementary insight into domain-specific topic interpretability.

\bibliographystyle{acl_natbib}
\bibliography{anthology,custom}

@inproceedings{han-etal-2023-unified,
    title = "Unified Neural Topic Model via Contrastive Learning and Term Weighting",
    author = "Han, Sungwon  and
      Shin, Mingi  and
      Park, Sungkyu  and
      Jung, Changwook  and
      Cha, Meeyoung",
    editor = "Vlachos, Andreas  and
      Augenstein, Isabelle",
    booktitle = "Proceedings of the 17th Conference of the European Chapter of the Association for Computational Linguistics",
    month = may,
    year = "2023",
    address = "Dubrovnik, Croatia",
    publisher = "Association for Computational Linguistics",
    url = "https://aclanthology.org/2023.eacl-main.132/",
    doi = "10.18653/v1/2023.eacl-main.132",
    pages = "1802--1817"
}

@inproceedings{bianchi-etal-2021-cross,
    title = "Cross-lingual Contextualized Topic Models with Zero-shot Learning",
    author = "Bianchi, Federico  and
      Terragni, Silvia  and
      Hovy, Dirk  and
      Nozza, Debora  and
      Fersini, Elisabetta",
    editor = "Merlo, Paola  and
      Tiedemann, Jorg  and
      Tsarfaty, Reut",
    booktitle = "Proceedings of the 16th Conference of the European Chapter of the Association for Computational Linguistics: Main Volume",
    month = apr,
    year = "2021",
    address = "Online",
    publisher = "Association for Computational Linguistics",
    url = "https://aclanthology.org/2021.eacl-main.143/",
    doi = "10.18653/v1/2021.eacl-main.143",
    pages = "1676--1683"
}

@inproceedings{akash-chang-2024-enhancing,
    title = "Enhancing Short-Text Topic Modeling with {LLM}-Driven Context Expansion and Prefix-Tuned {VAE}s",
    author = "Akash, Pritom Saha  and
      Chang, Kevin Chen-Chuan",
    editor = "Al-Onaizan, Yaser  and
      Bansal, Mohit  and
      Chen, Yun-Nung",
    booktitle = "Findings of the Association for Computational Linguistics: EMNLP 2024",
    month = nov,
    year = "2024",
    address = "Miami, Florida, USA",
    publisher = "Association for Computational Linguistics",
    url = "https://aclanthology.org/2024.findings-emnlp.917/",
    doi = "10.18653/v1/2024.findings-emnlp.917",
    pages = "15635--15646"
}

@inproceedings{adhya-sanyal-2024-ginopic,
    title = "{GIN}opic: Topic Modeling with Graph Isomorphism Network",
    author = "Adhya, Suman  and
      Sanyal, Debarshi Kumar",
    editor = "Duh, Kevin  and
      Gomez, Helena  and
      Bethard, Steven",
    booktitle = "Proceedings of the 2024 Conference of the North American Chapter of the Association for Computational Linguistics: Human Language Technologies (Volume 1: Long Papers)",
    month = jun,
    year = "2024",
    address = "Mexico City, Mexico",
    publisher = "Association for Computational Linguistics",
    url = "https://aclanthology.org/2024.naacl-long.342/",
    doi = "10.18653/v1/2024.naacl-long.342",
    pages = "6171--6183"
}

@inproceedings{liu-etal-2025-neural,
    title = "Neural Topic Modeling via Contextual and Graph Information Fusion",
    author = "Liu, Jiyuan  and
      Yan, Jiaxing  and
      Zhu, Chunjiang  and
      Liu, Xingyu  and
      Qing, Li  and
      Rao, Yanghui",
    editor = "Christodoulopoulos, Christos  and
      Chakraborty, Tanmoy  and
      Rose, Carolyn  and
      Peng, Violet",
    booktitle = "Proceedings of the 2025 Conference on Empirical Methods in Natural Language Processing",
    month = nov,
    year = "2025",
    address = "Suzhou, China",
    publisher = "Association for Computational Linguistics",
    url = "https://aclanthology.org/2025.emnlp-main.670/",
    doi = "10.18653/v1/2025.emnlp-main.670",
    pages = "13247--13263",
    ISBN = "979-8-89176-332-6"
}

@inproceedings{pham-etal-2024-neuromax,
    title = "{N}euro{M}ax: Enhancing Neural Topic Modeling via Maximizing Mutual Information and Group Topic Regularization",
    author = "Pham, Duy-Tung  and
      Nguyen Vu, Thien Trang  and
      Nguyen, Tung  and
      Ngo, Linh Van  and
      Nguyen, Duc Anh  and
      Nguyen, Thien Huu",
    editor = "Al-Onaizan, Yaser  and
      Bansal, Mohit  and
      Chen, Yun-Nung",
    booktitle = "Findings of the Association for Computational Linguistics: EMNLP 2024",
    month = nov,
    year = "2024",
    address = "Miami, Florida, USA",
    publisher = "Association for Computational Linguistics",
    url = "https://aclanthology.org/2024.findings-emnlp.457/",
    doi = "10.18653/v1/2024.findings-emnlp.457",
    pages = "7758--7772"
}

@inproceedings{wang-etal-2021-cross-lingual,
    title = "Cross-lingual Text Classification with Heterogeneous Graph Neural Network",
    author = "Wang, Ziyun  and
      Liu, Xuan  and
      Yang, Peiji  and
      Liu, Shixing  and
      Wang, Zhisheng",
    editor = "Zong, Chengqing  and
      Xia, Fei  and
      Li, Wenjie  and
      Navigli, Roberto",
    booktitle = "Proceedings of the 59th Annual Meeting of the Association for Computational Linguistics and the 11th International Joint Conference on Natural Language Processing (Volume 2: Short Papers)",
    month = aug,
    year = "2021",
    address = "Online",
    publisher = "Association for Computational Linguistics",
    url = "https://aclanthology.org/2021.acl-short.78/",
    doi = "10.18653/v1/2021.acl-short.78",
    pages = "612--620"
}

@inproceedings{zhu-etal-2018-graphbtm,
    title = "{G}raph{BTM}: Graph Enhanced Autoencoded Variational Inference for Biterm Topic Model",
    author = "Zhu, Qile  and
      Feng, Zheng  and
      Li, Xiaolin",
    editor = "Riloff, Ellen  and
      Chiang, David  and
      Hockenmaier, Julia  and
      Tsujii, Jun{'}ichi",
    booktitle = "Proceedings of the 2018 Conference on Empirical Methods in Natural Language Processing",
    month = oct # "-" # nov,
    year = "2018",
    address = "Brussels, Belgium",
    publisher = "Association for Computational Linguistics",
    url = "https://aclanthology.org/D18-1495/",
    doi = "10.18653/v1/D18-1495",
    pages = "4663--4672"
}

@inproceedings{zhou-etal-2020-neural,
    title = "Neural Topic Modeling by Incorporating Document Relationship Graph",
    author = "Zhou, Deyu  and
      Hu, Xuemeng  and
      Wang, Rui",
    editor = "Webber, Bonnie  and
      Cohn, Trevor  and
      He, Yulan  and
      Liu, Yang",
    booktitle = "Proceedings of the 2020 Conference on Empirical Methods in Natural Language Processing (EMNLP)",
    month = nov,
    year = "2020",
    address = "Online",
    publisher = "Association for Computational Linguistics",
    url = "https://aclanthology.org/2020.emnlp-main.310/",
    doi = "10.18653/v1/2020.emnlp-main.310",
    pages = "3790--3796"
}

@inproceedings{pennington-etal-2014-glove,
    title = "{G}lo{V}e: Global Vectors for Word Representation",
    author = "Pennington, Jeffrey  and
      Socher, Richard  and
      Manning, Christopher",
    editor = "Moschitti, Alessandro  and
      Pang, Bo  and
      Daelemans, Walter",
    booktitle = "Proceedings of the 2014 Conference on Empirical Methods in Natural Language Processing ({EMNLP})",
    month = oct,
    year = "2014",
    address = "Doha, Qatar",
    publisher = "Association for Computational Linguistics",
    url = "https://aclanthology.org/D14-1162/",
    doi = "10.3115/v1/D14-1162",
    pages = "1532--1543"
}

@inproceedings{li-liang-2021-prefix,
    title = "Prefix-Tuning: Optimizing Continuous Prompts for Generation",
    author = "Li, Xiang Lisa  and
      Liang, Percy",
    editor = "Zong, Chengqing  and
      Xia, Fei  and
      Li, Wenjie  and
      Navigli, Roberto",
    booktitle = "Proceedings of the 59th Annual Meeting of the Association for Computational Linguistics and the 11th International Joint Conference on Natural Language Processing (Volume 1: Long Papers)",
    month = aug,
    year = "2021",
    address = "Online",
    publisher = "Association for Computational Linguistics",
    url = "https://aclanthology.org/2021.acl-long.353/",
    doi = "10.18653/v1/2021.acl-long.353",
    pages = "4582--4597"
}

@inproceedings{bianchi-etal-2021-pre,
    title = "Pre-training is a Hot Topic: Contextualized Document Embeddings Improve Topic Coherence",
    author = "Bianchi, Federico  and
      Terragni, Silvia  and
      Hovy, Dirk",
    editor = "Zong, Chengqing  and
      Xia, Fei  and
      Li, Wenjie  and
      Navigli, Roberto",
    booktitle = "Proceedings of the 59th Annual Meeting of the Association for Computational Linguistics and the 11th International Joint Conference on Natural Language Processing (Volume 2: Short Papers)",
    month = aug,
    year = "2021",
    address = "Online",
    publisher = "Association for Computational Linguistics",
    url = "https://aclanthology.org/2021.acl-short.96/",
    doi = "10.18653/v1/2021.acl-short.96",
    pages = "759--766"
}

@inproceedings{song-etal-2021-fast,
    title = "Fast {W}ord{P}iece Tokenization",
    author = "Song, Xinying  and
      Salcianu, Alex  and
      Song, Yang  and
      Dopson, Dave  and
      Zhou, Denny",
    editor = "Moens, Marie-Francine  and
      Huang, Xuanjing  and
      Specia, Lucia  and
      Yih, Scott Wen-tau",
    booktitle = "Proceedings of the 2021 Conference on Empirical Methods in Natural Language Processing",
    month = nov,
    year = "2021",
    address = "Online and Punta Cana, Dominican Republic",
    publisher = "Association for Computational Linguistics",
    url = "https://aclanthology.org/2021.emnlp-main.160/",
    doi = "10.18653/v1/2021.emnlp-main.160",
    pages = "2089--2103"
}

@inproceedings{wu-etal-2024-towards-topmost,
    title = "Towards the {T}op{M}ost: A Topic Modeling System Toolkit",
    author = "Wu, Xiaobao  and
      Pan, Fengjun  and
      Luu, Anh Tuan",
    editor = "Cao, Yixin  and
      Feng, Yang  and
      Xiong, Deyi",
    booktitle = "Proceedings of the 62nd Annual Meeting of the Association for Computational Linguistics (Volume 3: System Demonstrations)",
    month = aug,
    year = "2024",
    address = "Bangkok, Thailand",
    publisher = "Association for Computational Linguistics",
    url = "https://aclanthology.org/2024.acl-demos.4/",
    doi = "10.18653/v1/2024.acl-demos.4",
    pages = "31--41"
}

@inproceedings{angelov-inkpen-2024-topic,
    title = "Topic Modeling: Contextual Token Embeddings Are All You Need",
    author = "Angelov, Dimo  and
      Inkpen, Diana",
    editor = "Al-Onaizan, Yaser  and
      Bansal, Mohit  and
      Chen, Yun-Nung",
    booktitle = "Findings of the Association for Computational Linguistics: EMNLP 2024",
    month = nov,
    year = "2024",
    address = "Miami, Florida, USA",
    publisher = "Association for Computational Linguistics",
    url = "https://aclanthology.org/2024.findings-emnlp.790/",
    doi = "10.18653/v1/2024.findings-emnlp.790",
    pages = "13528--13539"
}

@inproceedings{lau-etal-2014-machine,
    title = "Machine Reading Tea Leaves: Automatically Evaluating Topic Coherence and Topic Model Quality",
    author = "Lau, Jey Han  and
      Newman, David  and
      Baldwin, Timothy",
    editor = "Wintner, Shuly  and
      Goldwater, Sharon  and
      Riezler, Stefan",
    booktitle = "Proceedings of the 14th Conference of the {E}uropean Chapter of the Association for Computational Linguistics",
    month = apr,
    year = "2014",
    address = "Gothenburg, Sweden",
    publisher = "Association for Computational Linguistics",
    url = "https://aclanthology.org/E14-1056/",
    doi = "10.3115/v1/E14-1056",
    pages = "530--539"
}

@inproceedings{nan-etal-2019-topic,
    title = "Topic Modeling with {W}asserstein Autoencoders",
    author = "Nan, Feng  and
      Ding, Ran  and
      Nallapati, Ramesh  and
      Xiang, Bing",
    editor = "Korhonen, Anna  and
      Traum, David  and
      M{\`a}rquez, Llu{\'i}s",
    booktitle = "Proceedings of the 57th Annual Meeting of the Association for Computational Linguistics",
    month = jul,
    year = "2019",
    address = "Florence, Italy",
    publisher = "Association for Computational Linguistics",
    url = "https://aclanthology.org/P19-1640/",
    doi = "10.18653/v1/P19-1640",
    pages = "6345--6381"
}

@article{dieng-etal-2020-topic,
    title = "Topic Modeling in Embedding Spaces",
    author = "Dieng, Adji B.  and
      Ruiz, Francisco J. R.  and
      Blei, David M.",
    editor = "Johnson, Mark  and
      Roark, Brian  and
      Nenkova, Ani",
    journal = "Transactions of the Association for Computational Linguistics",
    volume = "8",
    year = "2020",
    address = "Cambridge, MA",
    publisher = "MIT Press",
    url = "https://aclanthology.org/2020.tacl-1.29/",
    doi = "10.1162/tacl_a_00325",
    pages = "439--453"
}

@inproceedings{godey-etal-2024-anisotropy,
    title = "Anisotropy Is Inherent to Self-Attention in Transformers",
    author = "Godey, Nathan  and
      Clergerie, {\'E}ric  and
      Sagot, Beno{\^i}t",
    editor = "Graham, Yvette  and
      Purver, Matthew",
    booktitle = "Proceedings of the 18th Conference of the European Chapter of the Association for Computational Linguistics (Volume 1: Long Papers)",
    month = mar,
    year = "2024",
    address = "St. Julian{'}s, Malta",
    publisher = "Association for Computational Linguistics",
    url = "https://aclanthology.org/2024.eacl-long.3/",
    doi = "10.18653/v1/2024.eacl-long.3",
    pages = "35--48"
}

@inproceedings{li-etal-2025-topic,
    title = "Topic-Guided Reinforcement Learning with {LLM}s for Enhancing Multi-Document Summarization",
    author = "Li, Chuyuan  and
      Xu, Austin  and
      Joty, Shafiq  and
      Carenini, Giuseppe",
    editor = "Christodoulopoulos, Christos  and
      Chakraborty, Tanmoy  and
      Rose, Carolyn  and
      Peng, Violet",
    booktitle = "Findings of the Association for Computational Linguistics: EMNLP 2025",
    month = nov,
    year = "2025",
    address = "Suzhou, China",
    publisher = "Association for Computational Linguistics",
    url = "https://aclanthology.org/2025.findings-emnlp.662/",
    doi = "10.18653/v1/2025.findings-emnlp.662",
    pages = "12395--12412",
    ISBN = "979-8-89176-335-7"
}

@article{blei2003latent,
  title={Latent dirichlet allocation},
  author={Blei, David M and Ng, Andrew Y and Jordan, Michael I},
  journal={Journal of machine Learning research},
  volume={3},
  number={Jan},
  pages={993--1022},
  year={2003}
}

@inproceedings{miao2016neural,
  title={Neural variational inference for text processing},
  author={Miao, Yishu and Yu, Lei and Blunsom, Phil},
  booktitle={International conference on machine learning},
  pages={1727--1736},
  year={2016},
  organization={PMLR}
}

@inproceedings{srivastava2017autoencoding,
  title={Autoencoding variational inference for topic models},
  author={Srivastava, Akash and Sutton, Charles},
  booktitle={International Conference on Learning Representations},
  year={2017}
}

@inproceedings{wu2024fastopic,
  title={FASTopic: A Fast, Adaptive, Stable, and Transferable Topic Modeling Paradigm},
  author={Wu, Xiaobao and Nguyen, Thong and Zhang, Delvin and Wang, William Yang and Luu, Anh Tuan},
  booktitle={Advances in Neural Information Processing Systems},
  volume={37},
  year={2024}
}

@article{kingma2013auto,
  title={Auto-encoding variational bayes},
  author={Kingma, Diederik P and Welling, Max},
  journal={arXiv preprint arXiv:1312.6114},
  year={2013}
}

@misc{grootendorst2022bertopicneuraltopicmodeling,
      title={BERTopic: Neural topic modeling with a class-based TF-IDF procedure}, 
      author={Maarten Grootendorst},
      year={2022},
      eprint={2203.05794},
      archivePrefix={arXiv},
      primaryClass={cs.CL},
      url={https://arxiv.org/abs/2203.05794}, 
}

@inproceedings{donabauer2025token,
  title={Token-Level Graphs for Short Text Classification},
  author={Donabauer, Gregor and Kruschwitz, Udo},
  booktitle={European Conference on Information Retrieval},
  pages={427--436},
  year={2025},
  organization={Springer}
}

@inproceedings{reimers-gurevych-2019-sentence,
    title = "Sentence-{BERT}: Sentence Embeddings using {S}iamese 
              {BERT}-Networks",
    author = "Reimers, Nils and Gurevych, Iryna",
    booktitle = "Proceedings of the 2019 Conference on Empirical Methods 
                 in Natural Language Processing and the 9th International 
                 Joint Conference on Natural Language Processing (EMNLP-IJCNLP)",
    month = nov,
    year = "2019",
    address = "Hong Kong, China",
    publisher = "Association for Computational Linguistics",
    pages = "3982--3992",
}

@article{wu2023graph,
  title={Graph neural networks for natural language processing: A survey},
  author={Wu, Lingfei and Chen, Yu and Shen, Kai and Guo, Xiaojie and Gao, Hanning and Li, Shucheng and Pei, Jian and Long, Bo},
  journal={Foundations and Trends in Machine Learning},
  volume={16},
  number={2},
  pages={119--328},
  year={2023},
  publisher={Emerald Publishing Limited}
}

@ARTICLE{6778764,
  author={Cheng, Xueqi and Yan, Xiaohui and Lan, Yanyan and Guo, Jiafeng},
  journal={IEEE Transactions on Knowledge and Data Engineering}, 
  title={BTM: Topic Modeling over Short Texts}, 
  year={2014},
  volume={26},
  number={12},
  pages={2928-2941},
  doi={10.1109/TKDE.2014.2313872}}

@inproceedings{
xu2023contextguided,
title={Context-guided Embedding Adaptation for Effective Topic Modeling in Low-Resource Regimes},
author={Yishi Xu and Jianqiao Sun and Yudi Su and Xinyang Liu and Zhibin Duan and Bo Chen and Mingyuan Zhou},
booktitle={Thirty-seventh Conference on Neural Information Processing Systems},
year={2023},
url={https://openreview.net/forum?id=cYkSt7jqlx}
}

@article{wang2024graph,
  title={Graph neural networks for text classification: A survey},
  author={Wang, Kunze and Ding, Yihao and Han, Soyeon Caren},
  journal={Artificial intelligence review},
  volume={57},
  number={8},
  pages={190},
  year={2024},
  publisher={Springer}
}

@inproceedings{
kipf2017semisupervised,
title={Semi-Supervised Classification with Graph Convolutional Networks},
author={Thomas N. Kipf and Max Welling},
booktitle={International Conference on Learning Representations},
year={2017},
url={https://openreview.net/forum?id=SJU4ayYgl}
}

@inproceedings{wang-etal-2021-extracting,
  title     = {Extracting Topics with Simultaneous Word Co-occurrence 
               and Semantic Correlation Graphs: Neural Topic Modeling 
               for Short Texts},
  author    = {Wang, Yiming and Li, Ximing and Zhou, Xiaotang and 
               Ouyang, Jihong},
  booktitle = {Findings of the Association for Computational 
               Linguistics: EMNLP 2021},
  pages     = {18--27},
  year      = {2021}
}

@article{lee-etal-2020-biobert,
  title     = {{BioBERT}: A Pre-trained Biomedical Language 
               Representation Model for Biomedical Text Mining},
  author    = {Lee, Jinhyuk and Yoon, Wonjin and Kim, Sungdong and 
               Kim, Donghyeon and Kim, Sunkyu and So, Chan Ho and 
               Kang, Jaewoo},
  journal   = {Bioinformatics},
  volume    = {36},
  number    = {4},
  pages     = {1234--1240},
  year      = {2020},
  publisher = {Oxford University Press}
}

@inproceedings{kornilova2019billsum,
  title={BillSum: A corpus for automatic summarization of US legislation},
  author={Kornilova, Anastassia and Eidelman, Vladimir},
  booktitle={Proceedings of the 2nd Workshop on New Frontiers in Summarization},
  pages={48--56},
  year={2019}
}

@article{wu2024survey,
  title={A survey on neural topic models: methods, applications, and challenges},
  author={Wu, Xiaobao and Nguyen, Thong and Luu, Anh Tuan},
  journal={Artificial Intelligence Review},
  volume={57},
  number={2},
  pages={18},
  year={2024},
  publisher={Springer}
}

@inproceedings{wu2023effective,
  title={Effective neural topic modeling with embedding clustering regularization},
  author={Wu, Xiaobao and Dong, Xinshuai and Nguyen, Thong Thanh and Luu, Anh Tuan},
  booktitle={International Conference on Machine Learning},
  pages={37335--37357},
  year={2023},
  organization={PMLR}
}

@book{Manning_IR_2008, place={Cambridge}, title={Introduction to Information Retrieval}, publisher={Cambridge University Press}, author={Manning, Christopher D. and Raghavan, Prabhakar and Schütze, Hinrich}, year={2008}}

@article{kingma2014adam,
  title={Adam: A method for stochastic optimization},
  author={Kingma, Diederik P and Ba, Jimmy},
  journal={arXiv preprint arXiv:1412.6980},
  year={2014}
}

@inproceedings{
gao2018representation,
title={Representation Degeneration Problem in Training Natural Language Generation Models},
author={Jun Gao and Di He and Xu Tan and Tao Qin and Liwei Wang and Tieyan Liu},
booktitle={International Conference on Learning Representations},
year={2019},
url={https://openreview.net/forum?id=SkEYojRqtm},
}

@article{hofmann2013probabilistic,
  title={Probabilistic latent semantic analysis},
  author={Hofmann, Thomas},
  journal={arXiv preprint arXiv:1301.6705},
  year={2013}
}

@article{mikolov2013efficient,
  title={Efficient estimation of word representations in vector space},
  author={Mikolov, Tomas and Chen, Kai and Corrado, Greg and Dean, Jeffrey},
  journal={arXiv preprint arXiv:1301.3781},
  year={2013}
}

@ARTICLE{9112648,
  author={Gui, Lin and Leng, Jia and Zhou, Jiyun and Xu, Ruifeng and He, Yulan},
  journal={IEEE Transactions on Knowledge and Data Engineering}, 
  title={Multi Task Mutual Learning for Joint Sentiment Classification and Topic Detection}, 
  year={2022},
  volume={34},
  number={4},
  pages={1915-1927},
  doi={10.1109/TKDE.2020.2999489}}

@inproceedings{10.1145/3477495.3531990,
author = {Zhang, Yuxiang and Jiang, Tao and Yang, Tianyu and Li, Xiaoli and Wang, Suge},
title = {HTKG: Deep Keyphrase Generation with Neural Hierarchical Topic Guidance},
year = {2022},
isbn = {9781450387323},
publisher = {Association for Computing Machinery},
address = {New York, NY, USA},
url = {https://doi.org/10.1145/3477495.3531990},
doi = {10.1145/3477495.3531990},
booktitle = {Proceedings of the 45th International ACM SIGIR Conference on Research and Development in Information Retrieval},
pages = {1044–1054},
numpages = {11},
location = {Madrid, Spain},
series = {SIGIR '22}
}
\appendix

\section{Benchmark Datasets}
\label{sec:appendix dataset}

In this section, we provide detailed description of the benchmark datasets: 20NewsGroup\footnote{\url{https://github.com/MIND-Lab/OCTIS/tree/master/preprocessed_datasets/20NewsGroup}}, BioASQ\footnote{\url{https://github.com/AdhyaSuman/GINopic/tree/master/preprocessed_datasets/Bio}} and Bills\footnote{\url{https://huggingface.co/datasets/FiscalNote/billsum}}. The statistics of the preprocessed datasets are presented in Table \ref{tab:dataset_statistics}.

\begin{table*}[t]
\centering
\begin{tabular}{l l c c c c}
\hline
\textbf{Dataset} & \textbf{Domain} & \textbf{\#Docs} & \textbf{\#Vocab} & \textbf{Avg. length} & \textbf{\#Labels} \\ \hline
\textbf{20NewsGroup} & \textit{General}     & 16,309 & 1,612 & 48.02 & 20 \\
\textbf{BioASQ}        & \textit{Biomedical}  & 19,448 & 2,000 & 7.44 & 20 \\
\textbf{Bills}         & \textit{Legal}       & 18,945  & 2,000 & 76.28 & -- \\ \hline
\end{tabular}
\caption{Statistics of the preprocessed benchmark datasets}
\label{tab:dataset_statistics}
\end{table*}

\section{Implementation Details and Hyperparameter Settings}
\label{sec:implementation}

In this section, we describe the training environment, our proposed \textsc{DARTopic} framework and baseline models details.
All experiments, including those for \textsc{DARTopic} and all baseline methods, were conducted under the same training environment for fair comparison. Specifically, all models were implemented in PyTorch (version 2.10.0) and trained on a single NVIDIA H100 GPU. Following a unified setup, we used \texttt{all-MiniLM-L6-v2}\footnote{\url{https://huggingface.co/sentence-transformers/all-MiniLM-L6-v2}} as the PLM for all methods, and adopted Adam \citep{kingma2014adam} as the optimizer with a learning rate of $1\times10^{-3}$ for 200 epochs.
For \textbf{FASTopic}, we set $\mathrm{DT}_{\alpha}=3.0$, $\mathrm{TW}_{\alpha}=2.0$, and $\theta_{\mathrm{temp}}=1.0$.
For \textbf{NeuroMax}, we followed the hyperparameter settings provided in the official implementation. In particular, we used $\epsilon=1\times10^{-16}$, $\mathrm{OT\_max\_iter}=5000$, $\mathrm{stopThr}=0.5\times10^{-2}$, $\mathrm{num\_groups}=10$, $\beta_{\mathrm{temp}}=0.2$, $\mathrm{weight\_loss\_ECR}=250.0$, $\mathrm{weight\_loss\_GR}=250.0$, $\alpha_{\mathrm{GR}}=20.0$, $\alpha_{\mathrm{ECR}}=20.0$, $\mathrm{sinkhorn\_max\_iter}=1000$, and $\mathrm{weight\_loss\_InfoNCE}=10.0$.
For \textbf{GINopic}, we used pre-trained Word2Vec\footnote{\url{https://huggingface.co/fse/word2vec-google-news-300}} embeddings for graph construction. The edge-weight threshold was set to $0.4$ for Bills and 20NG, and $0.05$ for BioASQ.
For \textbf{CGTM}, we set $\gamma_c=1.0$, $\gamma_g=0.2$, and GMM weight equal to 20. For graph information, we used $\epsilon_g=0.004$.

For our proposed \textsc{DARTopic}, the edge-weight threshold ($\tau$) was set to $0.2$ for BioASQ and 20NG, and $0.3$ for Bills.

\section{Evaluation Metrics}
\label{sec:evaluation_metrics}

In this section, we explain the details of our used automatic evaluation metrics in subsection \ref{sec:experimental setup}. We implement NPMI and TU metrics based on TopMost \citep{wu-etal-2024-towards-topmost}, a comprehensive toolkit for comparing and optimizing topic modeling in various scenarios\footnote{\url{https://github.com/BobXWu/TopMost}}. We evaluate topic coherence using Normalized Pointwise Mutual Information (NPMI) \citep{lau-etal-2014-machine}. For each topic \(k\), let \(W_k = \{w_1, \dots, w_L\}\) denote the top \(L\) words. The NPMI between a pair of words \((w_i, w_j)\) is defined as:

\begin{equation}
\mathrm{NPMI}(w_i, w_j)
=
\frac{
\log \frac{P(w_i, w_j)}{P(w_i)P(w_j)}
}{
-\log P(w_i, w_j)
}.
\end{equation}

Here, \(P(w_i)\) and \(P(w_j)\) denote the marginal probabilities of words \(w_i\) and \(w_j\), respectively, and \(P(w_i, w_j)\) denotes their co-occurrence probability estimated from a reference corpus. The topic-level coherence score is obtained by averaging the NPMI values over all word pairs among the top \(L\) words:

\begin{equation}
\mathrm{NPMI}(k)
=
\frac{2}{L(L-1)}
\sum_{1 \le i < j \le L}
\mathrm{NPMI}(w_i, w_j).
\end{equation}

The final NPMI score is computed by averaging across all \(K\) topics:

\begin{equation}
\mathrm{NPMI}
=
\frac{1}{K}
\sum_{k=1}^{K}
\mathrm{NPMI}(k).
\end{equation}

A higher NPMI score indicates that the top words within a topic co-occur more frequently in the reference corpus, suggesting better semantic coherence. Additionally, we evaluate topic diversity using topic uniqueness (TU) \citep{nan-etal-2019-topic}. For each topic \(k\), the top \(L\) words are obtained by ranking the decoder topic-word weights \(\beta_k\) in descending order. Let \(\mathrm{cnt}(l,k)\) denote the number of topics whose top-\(L\) words contain the \(l\)-th top word of topic \(k\). Then, the uniqueness of topic \(k\) is defined as:

\begin{equation}
\mathrm{TU}(k) = \frac{1}{L} \sum_{l=1}^{L} \frac{1}{\mathrm{cnt}(l,k)}, \quad k = 1, \dots, K.
\end{equation}

The final TU score is computed by averaging over all \(K\) topics:

\begin{equation}
\mathrm{TU} = \frac{1}{K} \sum_{k=1}^{K} \mathrm{TU}(k).
\end{equation}

The TU value ranges from \(1/K\) to \(1\), where a higher value indicates that fewer top words are shared across topics, implying greater topic diversity and lower redundancy. The overall topic quality (TQ) \citep{dieng-etal-2020-topic} is calculated as the product of the coherence and diversity values.

To evaluate the quality of document-topic distributions, we perform a document clustering task based on the inferred document-topic representations. The resulting clustering assignments are then evaluated using Purity and normalized mutual information (NMI) \citep{Manning_IR_2008}, which measure how well the predicted clusters align with the ground-truth document labels.

Let $\mathcal{C} = \{C_1, \dots, C_K\}$ denote the set of predicted clusters and $\mathcal{L} = \{L_1, \dots, L_J\}$ denote the set of ground-truth classes, where $N$ is the total number of documents. Purity measures the extent to which each cluster contains documents from a single class, and is defined as
\begin{equation}
\mathrm{Purity} = \frac{1}{N} \sum_{k=1}^{K} \max_{j} |C_k \cap L_j|.
\end{equation}
A higher Purity indicates that each predicted cluster is dominated by documents from one ground-truth class.

NMI measures the agreement between the predicted clustering and the ground-truth labels from an information-theoretic perspective. It is defined as
\begin{equation}
\mathrm{NMI}(\mathcal{C}, \mathcal{L}) =
\frac{I(\mathcal{C}; \mathcal{L})}
{\sqrt{H(\mathcal{C})H(\mathcal{L})}},
\end{equation}
where $I(\mathcal{C}; \mathcal{L})$ is the mutual information between the predicted clusters and the ground-truth classes, and $H(\mathcal{C})$ and $H(\mathcal{L})$ are their entropies. The mutual information is computed as
\begin{equation}
I(\mathcal{C}; \mathcal{L}) =
\sum_{k=1}^{K}\sum_{j=1}^{J}
\frac{|C_k \cap L_j|}{N}
\log
\frac{N \cdot |C_k \cap L_j|}{|C_k|\,|L_j|},
\end{equation}
and the entropy of the predicted clusters is defined as
\begin{equation}
H(\mathcal{C}) =
- \sum_{k=1}^{K} \frac{|C_k|}{N} \log \frac{|C_k|}{N},
\end{equation}
with $H(\mathcal{L})$ defined analogously for the ground-truth classes. Higher NMI indicates better consistency between the cluster assignments and the true labels.

\section{Analysis of Graph Construction Methods}
\label{sec:graph_contruction}

\begin{table}[H]
    \centering
    \resizebox{0.99\columnwidth}{!}{
    \begin{tabular}{lccc|ccc}
        \toprule
        \multirow{2}{*}{\centering Graph Structure} & \multicolumn{3}{c}{20NG} & \multicolumn{3}{c}{BioASQ} \\
        \cmidrule(lr){2-4} \cmidrule(lr){5-7}
        & NPMI & TU & TQ & NPMI & TU & TQ \\
        \midrule
        1-hop & 0.2643 & 0.9307 & 0.2460 & 0.1583 & 0.9600 & 0.1520 \\
        2-hop & 0.2667 & \textbf{0.9400} & 0.2507 & \textbf{0.1687} & 0.9600 & \textbf{0.1620} \\
        3-hop & 0.2655 & 0.9227 & 0.2449 & 0.1659 & 0.9693 & 0.1608 \\
        \midrule
        Semantic (ours) & \textbf{0.2736} & 0.9360 & \textbf{0.2561} & 0.1641 & \textbf{0.9760} & 0.1602 \\
        \bottomrule
    \end{tabular}}
    \caption{Analysis of different graph construction methods in our proposed \textsc{DARTopic} framework. The best-performing method is highlighted in \textbf{bold}.}
    \label{tab:graph_construction}
\end{table}

We explore two graph construction strategies, the token-level n-hop graph \citep{donabauer2025token} and the our token-level semantic graph method, within our \textsc{DARTopic} framework. In tabel \ref{tab:graph_construction}, the results show that the n-hop graph is more effective on BioASQ, which consists of relatively short documents, whereas the semantic graph performs better on 20NG, which contains relatively long documents. This suggests that local structural relations are more useful for short texts, while broader semantic connections become more important for longer documents. Overall, the results highlight that graph construction should be chosen carefully according to the document length characteristics of the target corpus.

\section{Hyperparameter Analysis}
We analyze the effect of the graph construction hyperparameter $\tau$ in DARTopic. The parameter $\tau$ controls the semantic similarity threshold used to construct the token-level semantic graph. To examine its influence, we evaluate $\tau \in \{0.1, 0.2, 0.3, 0.4\}$ on three datasets: 20NG, BioASQ, and Bills.

As shown in Table \ref{tab:hyperparameter_analysis}, the best topic quality is generally achieved when $\tau$ is set between 0.2 and 0.3. Specifically, 20NG obtains the highest TQ score when $\tau=0.2$, while Bills achieves the best TQ score when $\tau=0.3$. For BioASQ, the performance is also stable in this range, and we select $\tau=0.2$ for the main experiments. Thus, we use $\tau=0.2$ for 20NG and BioASQ, and $\tau=0.3$ for Bills.
The results show that both too small and too large values of $\tau$ can be less effective. When $\tau$ is too small, weak semantic relations may be included, which can introduce noisy edges into the graph. 
In contrast, when $\tau$ is too large, the graph may become overly sparse and discard useful semantic connections. 
Therefore, moderate values of $\tau$ provide a better balance between preserving meaningful semantic relations and filtering out noisy ones.

Interestingly, this tendency differs from n-hop based graph construction methods \citep{donabauer2025token}, whose performance can be highly affected by document length. Although BioASQ consists of relatively short documents, while 20NG and Bills contain longer documents, the optimal range of $\tau$ remains similar across all datasets. This indicates that the proposed token-level semantic graph construction is less sensitive to document length and less dependent on dataset-specific hyperparameter tuning. Overall, our analysis suggest that the proposed semantic graph construction method is robust to hyperparameter changes and can effectively adapt to datasets with different document lengths and domains.

\begin{table*}[t]
    \centering
    \resizebox{0.80\textwidth}{!}{
    \begin{tabular}{lccc|ccc|ccc}
        \toprule
        & \multicolumn{3}{c}{20NG} 
        & \multicolumn{3}{c}{BioASQ}
        & \multicolumn{3}{c}{Bills} \\
        \cmidrule(lr){2-4} \cmidrule(lr){5-7} \cmidrule(lr){8-10}
        & NPMI & TU & TQ
        & NPMI & TU & TQ
        & NPMI & TU & TQ \\
        \midrule
        $\tau$ = 0.1 & 0.2670 & 0.9320 & 0.2488 & 0.1668 & 0.9693 &	0.1617 & 0.2503	& 0.9307	& 0.2329 \\
        $\tau$ = 0.2 & 0.2736 &	0.9360	& 0.2561 & 0.1641 & 0.9760 &	0.1602 &  0.2545 &	0.9160	& 0.2331 \\
        $\tau$ = 0.3   & 0.2689 & 0.9307 &0.2502 & 0.1737 & 0.9533 & 0.1656 & 0.2685	& 0.9240	& 0.2481 \\
        $\tau$ = 0.4   & 0.2647	& 0.9400 & 0.2488 & 0.1609 & 0.9640 & 0.1551 & 0.2489	& 0.9173	& 0.2283 \\
        \bottomrule
    \end{tabular}}
    \caption{Analysis of graph construction hyperparameter in \textsc{DARTopic}.}
    \label{tab:hyperparameter_analysis}
\end{table*}

\section{Qualitative Evaluation}

In this section, we analyze to further examine the interpretability of the topics generated by different methods. 
Specifically, we select topics from the BioASQ dataset that contain biomedical-related keywords such as \textit{``acid'', ``dna'', ``rna''}, and \textit{``cell''}. 
For each method, we report two representative topic word lists in Table \ref{tab:qualitative_evaluation}. 
Words highlighted in \textcolor{blue}{blue} indicate general or less domain-specific terms that are relatively unrelated to professional biomedical terminology.

As shown in Table \ref{tab:qualitative_evaluation}, several baseline methods generate topics that include many general words, such as ``efficiency'', ``food'', ``human'', ``prediction'', ``experience'', and ``route''. 
Although these words may frequently appear in the corpus, they are less informative for explaining biomedical topics. 
This suggests that the baseline methods often mix domain-specific biomedical terms with general-purpose words, which can reduce topic interpretability.

In contrast, DARTopic produces more coherent and domain-specific topic words, including terms such as \textit{``replication'', ``leukemic'', ``rna'', ``avian'', ``leukemia'', ``adenovirus'', ``dna'', ``ribonucleic acid''}, and \textit{``messenger''}. These words are closely related to biomedical concepts and provide clearer semantic explanations of the learned topics. Therefore, the qualitative results demonstrate that DARTopic can generate more interpretable and specialized topics in domain-specific corpora such as BioASQ.

\begin{table*}[t]
\centering
\resizebox{0.99\textwidth}{!}{%
\begin{tabular}{l p{16cm}}
\toprule
\textbf{Methods} & \centering \textbf{Top-related word examples} \tabularnewline
\midrule
\multirow{2}{*}{ProdLDA} 
& \textcolor{blue}{efficiency} citrate acidosis \textcolor{blue}{african} \textcolor{blue}{section} diurnal gastrin diabetic  tube conformation \\
& \textcolor{blue}{food} guanethidine reagent hydroxy recognition testing acidosis trial \textcolor{blue}{dog} \textcolor{blue}{dark} \\
\midrule
\multirow{2}{*}{FASTopic} 
& \textcolor{blue}{human} cell \textcolor{blue}{skin} \textcolor{blue}{red} \textcolor{blue}{culture} erythrocytes peripheral serum \textcolor{blue}{studies} \textcolor{blue}{normal} \\
& acid synthesis amino acids enzyme phosphate biosynthesis ribonucleic rna dna \\
\midrule
\multirow{2}{*}{BERTopic} 
& cells \textcolor{blue}{human} cell cadmium zinc dna electron \textcolor{blue}{skin} \textcolor{blue}{culture} synthesis \\
& amino intestine intestinal acid albumin trypsin \textcolor{blue}{pig} \textcolor{blue}{pigs} \textcolor{blue}{small} chymotrypsin \\
\midrule
\multirow{2}{*}{ZeroShotTM} 
& plasmid recombination \textcolor{blue}{repair} dna coli excision rec replication \textcolor{blue}{defective} \textcolor{blue}{division} \\
& hela nucleic rna poliovirus histone simian simplex sendai herpesvirus \textcolor{blue}{newcastle} \\
\midrule
\multirow{2}{*}{CombinedTM} 
& \textcolor{blue}{mother} \textcolor{blue}{prediction} rna \textcolor{blue}{rest} virulence insufficiency tracheal sickle dynamic \textcolor{blue}{experience} \\
& \textcolor{blue}{range} \textcolor{blue}{mapping} \textcolor{blue}{map} intracellular \textcolor{blue}{head} occurrence rhesus angle \textcolor{blue}{eye} respiration \\
\midrule
\multirow{2}{*}{NeuroMax} 
& extracellular basal probe ontogeny \textcolor{blue}{grow} \textcolor{blue}{subject} \textcolor{blue}{modify} \textcolor{blue}{apply} react mediate \\
& rna \textcolor{blue}{person} recipient effector \textcolor{blue}{border} evolution sow kitten \textcolor{blue}{rhythm} \textcolor{blue}{route} \\
\midrule
\multirow{2}{*}{PVTM} 
& \textcolor{blue}{versus} hapten \textcolor{blue}{secondary} \textcolor{blue}{host} response cellular suppression \textcolor{blue}{class} hybrid \textcolor{blue}{carrier} \\
& nucleic hela rna ribonucleic poliovirus replication acid dna messenger  \textcolor{blue}{double} \\
\midrule
\multirow{2}{*}{GINopic} 
& \textcolor{blue}{spin} \textcolor{blue}{cross} rec dna \textcolor{blue}{head} \textcolor{blue}{description} viral inducible \textcolor{blue}{repair} hydroxybutyrate \\
& polymerase \textcolor{blue}{wild} dna hela \textcolor{blue}{freeze} synthase estradiol importance polarity malate \\
\midrule
\multirow{2}{*}{CGTM} 
& \textcolor{blue}{human} cell vitro peripheral surface antigen \textcolor{blue}{response} anti \textcolor{blue}{normal} \textcolor{blue}{culture} \\
& virus murine lymphocyte tumor \textcolor{blue}{mice} cell \textcolor{blue}{friend} spleen \textcolor{blue}{host} lymphocytic \\
\midrule
\multirow{2}{*}{\textsc{DARTopic}} 
& replication leukemic rauscher rna avian leukemia murine adenovirus primate dna \\
& ribonucleic acid aminoacyl ribosomal deoxyribonucleic nucleic trna synthetase messenger polynucleotide \\
\bottomrule
\end{tabular}%
}
\caption{Top-related word examples generated by different baseline methods.}
\label{tab:qualitative_evaluation}
\end{table*}

\section{Effect of Contextual Embeddings}

Word2Vec \citep{mikolov2013efficient} produces context-independent embeddings, edge weights are governed by pre-trained global word similarity rather than the actual semantic relationships within a document. This is the same word-level design adopted by GINopic, which falls substantially behind \textsc{DARTopic} across all three domains (Table \ref{tab:npmi-4datasets-acl-domain}). In contrast, our contextual token embeddings from a Transformer encoder (SBERT) allow the same token to obtain different embeddings depending on its co-occurring words. We also clarify that our goal is not to eliminate PLMs, but to reduce reliance on a specific PLM's scale or domain-specific pre-training (Tables \ref{tab:robustness_analysis_DARTopic} and \ref{tab:robustness_analysis_baselines}); the GCN layers compensate using corpus information. Nevertheless, we agree a Word2Vec-initialized variant is a valuable comparison and will add it to the revised version. As shown in Table \ref{tab:DARTopic_with_Word2Vec}, Word2Vec yields markedly lower topic quality in the biomedical domain, further underscoring the effectiveness of Transformer-based token embeddings.

\begin{table}[ht]
    \centering
    \resizebox{0.80\linewidth}{!}{%
    \begin{tabular}{lccc}
        \toprule
        & NPMI & TU & TQ \\
        \midrule
        \texttt{Word2Vec} & 0.0114 & 0.6093 & 0.0069 \\
        \texttt{MiniLM} & \textbf{0.1641} & \textbf{0.9760} & \textbf{0.1602} \\
        \bottomrule
    \end{tabular}%
    }
    \caption{Comparison of \textsc{DARTopic} performance on BioASQ dataset with Word2Vec models. The best-performing method is highlighted in \textbf{bold}.}
    \label{tab:DARTopic_with_Word2Vec}
\end{table}

\section{Use of AI Assistants}
AI assistants were used only for limited writing and language-editing support, as well as preparation of submission materials. All technical content, methodology, implementation, experimental results, and final verification were performed and checked by the authors.

\end{document}